\documentclass[conference]{IEEEtran}
\usepackage{times}
\usepackage{epsfig}
\usepackage{graphicx}
\usepackage{subcaption}
\usepackage{amsmath}
\usepackage{amssymb}
\usepackage{tikz}
\usepackage{multirow}
\usepackage{booktabs}
\usepackage{algorithm}
\usepackage{algpseudocode}
\usepackage{tabularx}
\usepackage{soul}
\usepackage{makecell}
\usepackage{caption}
\usepackage[pagebackref=true,breaklinks=true,letterpaper=true,colorlinks,bookmarks=false]{hyperref}

\definecolor{darkgreen}{rgb}{0, 0.5411, 0}

\IEEEoverridecommandlockouts

\newcommand{\printfnsymbol}[1]{%
  \textsuperscript{\@fnsymbol{#1}}%
}

\begin{document}

\title{Undercover Deepfakes: Detecting Fake Segments in Videos}
\author{
    \IEEEauthorblockN{
        Sanjay Saha\IEEEauthorrefmark{1}, 
        Rashindrie Perera\IEEEauthorrefmark{2}\textsuperscript{\textsection}, 
        Sachith Seneviratne\IEEEauthorrefmark{2}\textsuperscript{\textsection}, 
        Tamasha Malepathirana\IEEEauthorrefmark{2}, 
        Sanka Rasnayaka\IEEEauthorrefmark{1}, 
        \\Deshani Geethika\IEEEauthorrefmark{2}, 
        Terence Sim\IEEEauthorrefmark{1}, 
        Saman Halgamuge\IEEEauthorrefmark{2}
    }
    \IEEEauthorblockA{\IEEEauthorrefmark{1}National University of Singapore, Singapore}
    \IEEEauthorblockA{\IEEEauthorrefmark{2}University of Melbourne, Australia}
    \IEEEauthorblockA{
        sanjaysaha@u.nus.edu\IEEEauthorrefmark{1}, 
        cdperera@student.unimelb.edu.au\IEEEauthorrefmark{2},
        sachith.seneviratne@unimelb.edu.au\IEEEauthorrefmark{2}, 
        \\tmalepathira@student.unimelb.edu.au\IEEEauthorrefmark{2},
        sanka@nus.edu.sg\IEEEauthorrefmark{1} ,
        dpoddenige@student.unimelb.edu.au\IEEEauthorrefmark{2},
        \\terence.sim@nus.edu.sg\IEEEauthorrefmark{1},
        saman@unimelb.edu.au\IEEEauthorrefmark{2}
    }
}

\maketitle
\begingroup\renewcommand\thefootnote{\textsection}
\footnotetext{Equal contribution}
\endgroup
\thispagestyle{empty}

\begin{abstract}
The recent renaissance in generative models, driven primarily by the advent of diffusion models and iterative improvement in GAN methods, has enabled many creative applications. However, each advancement is also accompanied by a rise in the potential for misuse. In the arena of the deepfake generation, this is a key societal issue. In particular, the ability to modify segments of videos using such generative techniques creates a new paradigm of deepfakes which are mostly real videos altered slightly to distort the truth.
This paradigm has been under-explored by the current deepfake detection methods in the academic literature.
In this paper, we present a deepfake detection method that can address this issue by performing deepfake prediction at the frame and video levels. To facilitate testing our method, we prepared a new benchmark dataset where videos have both real and fake frame sequences with very subtle transitions. We provide a benchmark on the proposed dataset with our detection method which utilizes the Vision Transformer based on Scaling and Shifting \cite{lian2022scaling} to learn spatial features, and a Timeseries Transformer to learn temporal features of the videos to help facilitate the interpretation of possible deepfakes. Extensive experiments on a variety of deepfake generation methods show excellent results by the proposed method on temporal segmentation and classical video-level predictions as well. In particular, the paradigm we address will form a powerful tool for the moderation of deepfakes, where human oversight can be better targeted to the parts of videos suspected of being deepfakes. All experiments can be reproduced at: \href{https://github.com/sanjaysaha1311/temporal-deepfake-segmentation}{github.com/rgb91/temporal-deepfake-segmentation}.
\end{abstract}

\section{Introduction}

\begin{figure}[ht]
\centering
    \begin{subfigure}{0.49\textwidth}
        \includegraphics[width=\textwidth]{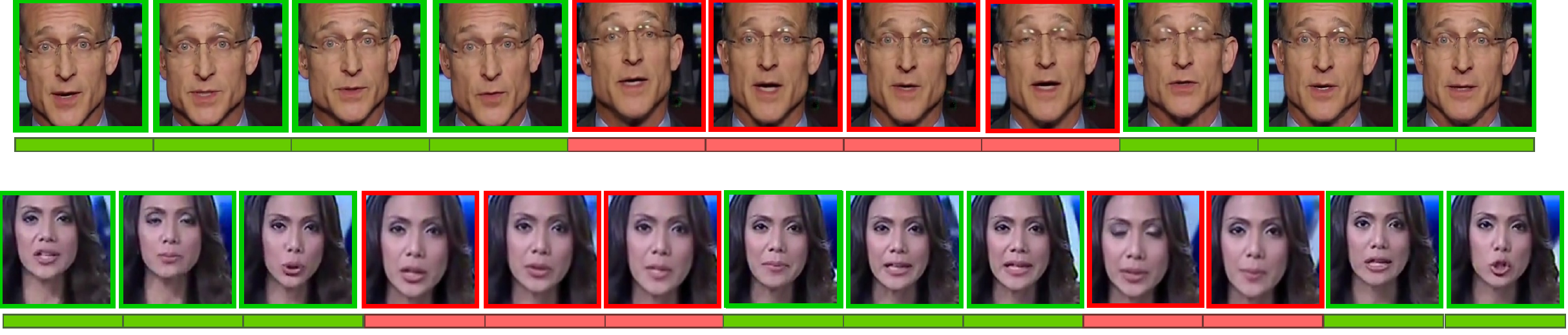}
        \caption{Simplified illustration of sample videos from the newly introduced benchmark dataset for temporal deepfake segment detection.}
        \vspace{2ex}
    \end{subfigure}
    \begin{subfigure}{0.49\textwidth}
        \includegraphics[width=\textwidth]{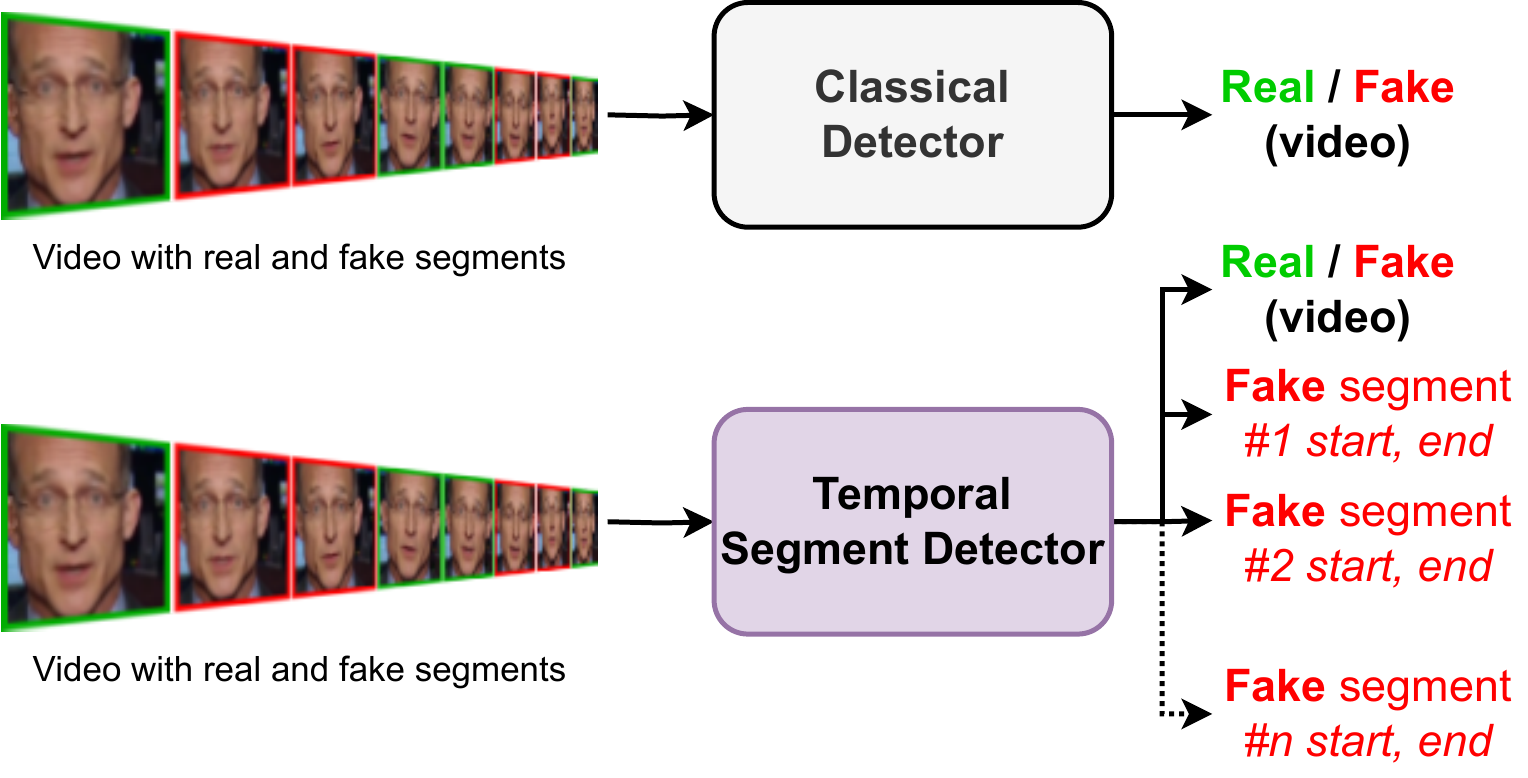}
        \caption{Comparison of our method with the classical deepfake detection.}
    \end{subfigure}
  \caption{(a) We propose a new deepfake benchmark dataset consisting of videos with one or two manipulated segments, represented in the two images respectively. Fake frames are indicated by smaller red boxes and genuine video frames are denoted by green borders. (b) Our proposed deepfake detection method employs temporal segmentation to classify video frames as real or fake and identify the intervals containing the manipulated content. This is a departure from the conventional binary classification of videos as either entirely genuine or entirely manipulated.}
  \label{fig:heading}
\end{figure}

Deep learning has made significant advances over the last few years, with varying degrees of societal impact. The advent of diffusion models as viable alternatives to hitherto established generative models has revolutionized the domains of textual/language learning, visual generation, and cross-modal transformations \cite{rombach2022high}. This, alongside other recent advancements in generative AI such as GPT4 \cite{bubeck2023sparks} has also brought to attention the social repercussions of advanced AI systems capable of realistic content generation.

There exist many methods to tackle the deepfake detection problem formulated as a binary classification problem \cite{afchar2018mesonet, bayar2016deep, cozzolino2017recasting}. A common pitfall of these methods is the inability to generalize to unseen deepfake creation methods. However, there is a more pressing drawback in these studies that we aim to highlight and address in this paper. Considering the social impact of a deepfake video, we hypothesize that rather than fabricating an entire fake video, a person with malicious intent would alter smaller portions of a video to misrepresent a person's views, ideology and public image. For example, an attacker can generate a few fake frames to replace some real frames in a political speech, thus distorting their political views which can lead to considerable controversy and infamy. 


The task of identifying deepfake alterations within a longer video, known as deepfake video temporal segmentation, is currently not well explored or understood. These types of deepfakes pose a more difficult challenge for automated deepfake analysis compared to other types of deepfakes. Moreover, they also pose a greater threat to society since the majority of the video may be legitimate, making it appear more realistic and convincing. Additionally, these deepfakes require significant manual oversight, especially in the moderation of online content platforms, as identifying the legitimacy of the entire video requires manual interpretation. However, performing frame-level detection allows the human moderator to save time by focusing only on the fake segments.

Figure \ref{fig:heading}(a) demonstrates deepfake videos where the entire video is not fake, but some of the real frames were replaced by fake frames. We present a benchmark dataset with videos similar to those in Figure \ref{fig:heading}(a) to test our method on the temporal deepfake segmentation problem. In the temporal deepfake segmentation problem the detector makes frame level predictions and calculates the start and end of the fake sequences i.e. fake-segments. This differs from classical deepfake detection where the detector makes a video level prediction as demonstrated in Figure \ref{fig:heading}(b).

\textbf{Problem Definition}

Deepfake Temporal Segmentation task is defined as,

\textit{Given an input video identify temporal segments within the video that are computer generated i.e. fakes.} The output of this task is a labeling of `real' or `fake' for each frame, which we call a temporal segmentation map. 
We can frame the classical deepfake detection problem as a special case of the temporal segmentation task, in which ALL frames are labeled either as `real' or `fake'. With an emphasis on the novel deepfake temporal segmentation task, this paper makes the following contributions,
\begin{itemize}
    \item We emphasize on the new threat of faking small parts of a longer video to pass it off as real. Current detection methods ignore this threat, since they assume the entire video is real or fake. This can be addressed through proposed temporal segmentation of videos. This provides a new direction for future research.
    \item We curated a new dataset specifically for deepfake temporal segmentation, which will be publicly available for researchers to evaluate their methods. Our rigorous experiments establish benchmark results for temporal segmentation of deepfakes, providing a baseline for future work.
\end{itemize}

\section{Related Work}
\textbf{Face Image Synthesis}:
Manipulation of face images has always been a popular research topic in the media forensics \cite{verdoliva2020media, rathgeb2022handbook} and biometrics domain. Synthesized digital faces can be used to deceive humans as well as machines and software. Prior to deepfakes, digitally manipulated faces \cite{realizable, sharif2016accessorize, korshunov2018deepfakes} were utilized mainly to fool biometric verification and identification methods e.g., face recognition systems. Consequently, deepfake methods \cite{stylegan, dfaker, fsgan, cyclegan} started to generate very realistic fake videos of faces and became much more popular as a result. This led to a series of research works on developing a number of deepfake generation methods, categorized into mainly two types: Face swapping \cite{deepfakes, faceshifter, deepfacelab} and Face reenactment \cite{face2face, neuraltexutres}. 

Deepfake generation methods have since improved significantly by advancing existing methods and better software integration, as in Deepfacelab \cite{deepfacelab}. This has helped creators of deepfakes to create longer videos, including seamlessly blending fake frames with real frames, which allows one to have both real and fake video segments within the same deepfake video. Through more recent developments in generative AI \cite{rombach2022high, tzaban2022stitch, nitzan2022mystyle} we are at the brink of experiencing even higher quality and more subtle deepfakes, raising the need for updated research in this area.

\textbf{Deepfake Detection}:
Initial works on deepfake detection methods \cite{stroebel2023systematic, rana2022deepfake, yu2021survey, passos2022review, ahmed2022analysis}  focused on detecting artifacts in deepfaked face images, such as irregular eye colors, asymmetric blinking eyes, abnormal heart beats, irregular lip, mouth and head movements \cite{li2018ictu, sun2021improving, ciftci2020fakecatcher, matern2019exploiting}. Some other earlier works tried to find higher-level variability in the videos: erroneous blending after face swaps, or identity-aware detection approach \cite{yang2019exposing, durall2019unmasking, facexray, cozzolino2021id}. Compared to these earlier works, more recent studies \cite{nguyen2019use, rahmouni2017distinguishing, amerini2019deepfake, zhuang2022uia, amoroso2023parents} that are independent of artifact-based detection have achieved astounding results in detecting fake videos from most of the state-of-the-art datasets. Recently, more works \cite{sladd, pairwise, madd, kim2021fretal, nadimpalli2022improving, korshunov2022improving, jain2022dataless, guan2023collaborative, zhao2021learning, shiohara2022detecting} have given increased attention towards generalizability of the detectors to detect deepfakes from unseen methods. 

\textbf{Temporal Segmentation}: Although deepfake detection methods have seen significant progress in recent years, only a few studies \cite{hernandez2022deepfakes, chugh2020not, cai2022you, he2021forgerynet} have looked into the problem of temporal segmentation task where in a long video only one or more short segments is altered while the rest of the frames are real. In this paper, we introduce a new, easily reproducible dataset based on the FaceForensics++ \cite{ffpp} and a method for not only detecting deepfake videos but also segmenting the fake frame-segments within them. The proposed approach can accurately identify one or more fake segments in a deepfake video, which can mitigate the risks associated with deepfakes that remain well-blended within real frames in a video.

\begin{figure*}[!ht]
  \includegraphics[width=\textwidth]{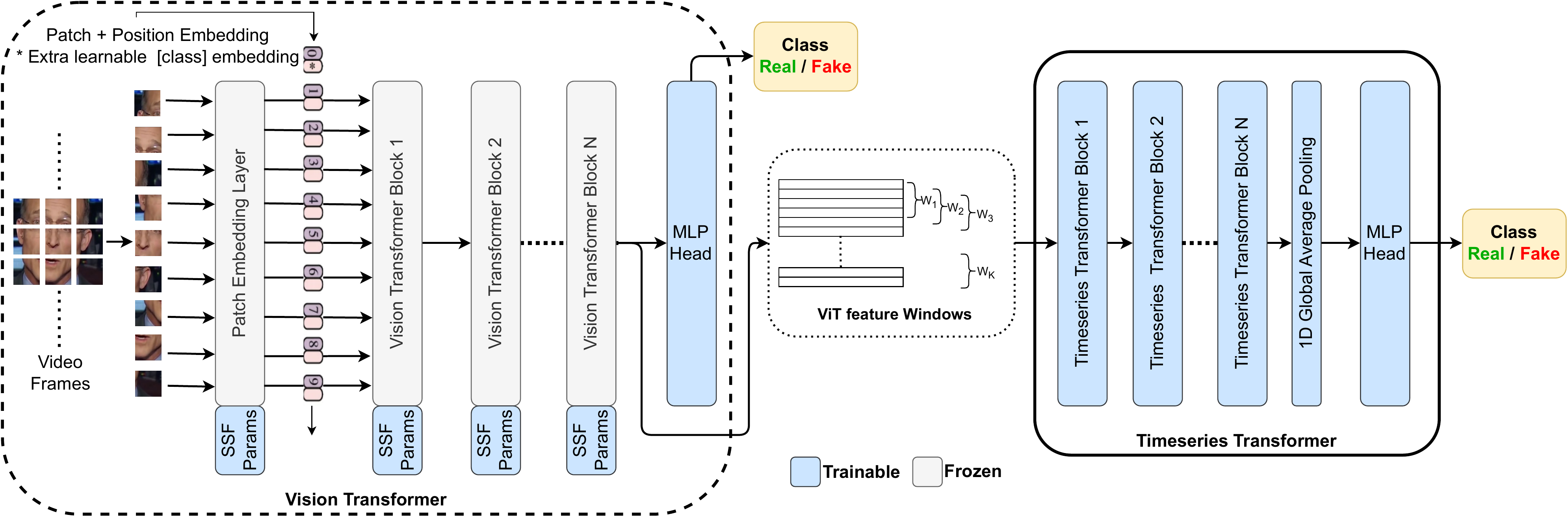}
  \caption{Our proposed detection method's model architecture comprises two main blocks: the Vision Transformer (ViT) and the Timeseries Transformer (TsT). The ViT fine-tunes a pre-trained model for deepfake detection using the Scaling and Shifting (SSF) method, learning spatial features. On the other hand, the TsT focuses on temporal features. The ViT's spatial features are sequentially accumulated and split into overlapping windows before inputting into the TsT.}
  \label{fig:archi}
\end{figure*}

\section{Methodology}

We propose a two-stage method as shown in Figure \ref{fig:archi}. The first stage employs a Vision Transformer (ViT) based on Scaling and Shifting (SSF) \cite{lian2022scaling}  to extract the frame-level features of the videos. Specifically, ViT learns a single vector representation for each frame and these feature vectors are sequentially accumulated and windowed using the sliding window technique.  The TsT is an adaptation of the original transformer encoder from \cite{attention}. TsT learns the temporal features from the learned ViT features and uses them for classification. The feature vectors from stage 1 are sequentially accumulated and windowed using the sliding window technique. It is important to use sequentially windowed feature vectors as input to the TsT since we need the temporal features to be learned for better temporal segmentation.

\subsection{Model architecture}

\subsubsection{Vision Transformer (ViT) and Scaling and Shifting (SSF)}

ViTs have achieved state-of-the-art results on several image classification benchmarks, demonstrating their effectiveness as an alternative to convolutional neural networks (CNNs). The employed ViT model first partitions the input image $I \in \Re ^{H\times W \times C}$ into a set of smaller patches of size $N \times N$ where $H$, $W$, $C$ and $N$ correspond to the height, width, number of channels of the image, and the height and width of each patch, respectively. Each patch is then represented by a $d$-dimensional feature vector, which is obtained by flattening the patch into a vector of size $N^2 C$ and applying a linear projection to reduce its dimensionality. Next, to allow the model to learn the spatial relationships between the patches, positional encodings are added to the patch embeddings. The resulting patch embeddings are concatenated together to form a sequence, and a learnable class embedding that represents the classification output is prepended to the sequence which is then input through a series of transformer layers. Each transformer layer consists of a multi-head self-attention mechanism, which allows the model to attend to different parts of the input patches, a multi-layer perceptron (MLP), and a layer normalization (Fig. \ref{fig:encoder_ssf}). Finally, a classification head is attached at the end of the transformer layers, which produces a probability distribution over the target classes.

Recently, there has been an upsurge in the use of parameter-efficient fine-tuning methods \cite{lian2022scaling, chen2022adaptformer} to fine-tune only a smaller subset of parameters in large pre-trained models such as ViTs, leading to better performance in downstream tasks compared to conventional end-to-end fine-tuning and linear probing. We use one such method, called SSF \cite{lian2022scaling} to fine-tune the pre-trained ViT model used in our pipeline (Fig. \ref{fig:archi}). SSF attempts to alleviate the distribution mismatch between the pre-trained task and the downstream deep fake feature extraction task by modulating deep features. Specifically, during the fine-tuning phase, the original network parameters are frozen, and SSF parameters are introduced at each operation to learn a linear transformation of the features, as shown in Fig. \ref{fig:encoder_ssf}. 

As done in the original work, we too insert SSF parameters after each operation including multi-head self-attention, MLP, layer normalization, etc. Specifically, given the input $x \in \Re^{N^2+1}\times d$, the output $y \in \Re^{N^2+1}\times d$ (also the input to the next operation) is calculated by
\vspace{-2mm}
\begin{equation}
y = \gamma \cdot x + \beta
\end{equation}
where $\gamma \in \Re^d$ and $\beta \in \Re^d$ are the scale and shift parameters, respectively.

\subsubsection{Timeseries Transformer (TsT) }\label{sec:method_timeseries}
\begin{figure}
\centering
    \begin{subfigure}{0.49\textwidth}
        \includegraphics[width=\textwidth]{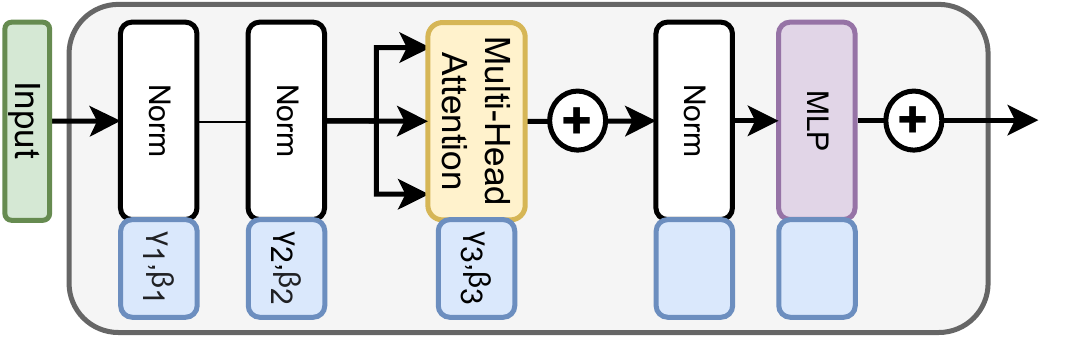}
        \caption{Vision Transformer Encoder}
        \vspace{2ex}
    \end{subfigure}
    \begin{subfigure}{0.18\textwidth}
        \includegraphics[width=0.95\textwidth]{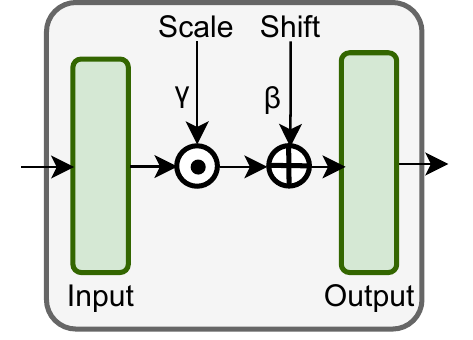}
        \caption{Scale and Shift}
    \end{subfigure}
    \begin{subfigure}{0.28\textwidth}
        \includegraphics[width=0.95\textwidth]{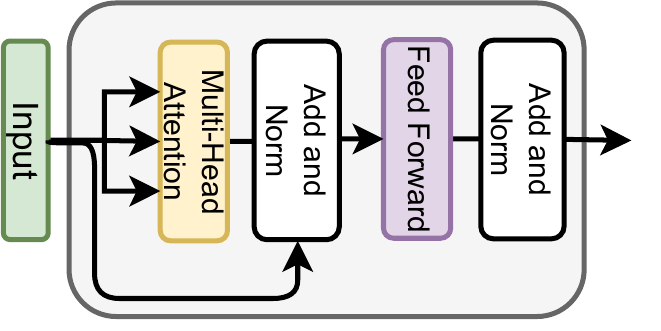}
        \caption{Timeseries Transformer Encoder}
    \end{subfigure}
    \caption{Architecture of the encoder blocks and Scale and Shift block.}
    \label{fig:encoder_ssf}
\end{figure}

\textbf{Architecture and training.} \quad The employed TsT is an adaptation of the sequence to sequence transformer in \cite{attention}. The transformer architecture is designed to learn and classify from sequential data instead of generating another sequence. It is composed of multiple transformer blocks and an MLP head. Each transformer block has a multi-head attention mechanism and a feed-forward block as shown in Figure \ref{fig:encoder_ssf}(c).

Our method generates frame-level predictions for the input videos, which may contain some noisy predictions. To address this issue, we used a simple smoothing technique based on majority voting over a sliding window of size $15$. It takes a majority vote from the predictions of the frames within the window around a particular frame. By smoothing out the noisy predictions, our approach improves performance, as demonstrated in Table \ref{tab:ablation}.

\textbf{Data processing.} \quad 
For the TsT, we accumulated the feature vectors from the ViT sequentially for each video and split them into overlapping windows of size $W$. That is, we have features of $W$ sequential frames in one window.

\begin{figure*}[ht]
    \includegraphics[width=\textwidth]{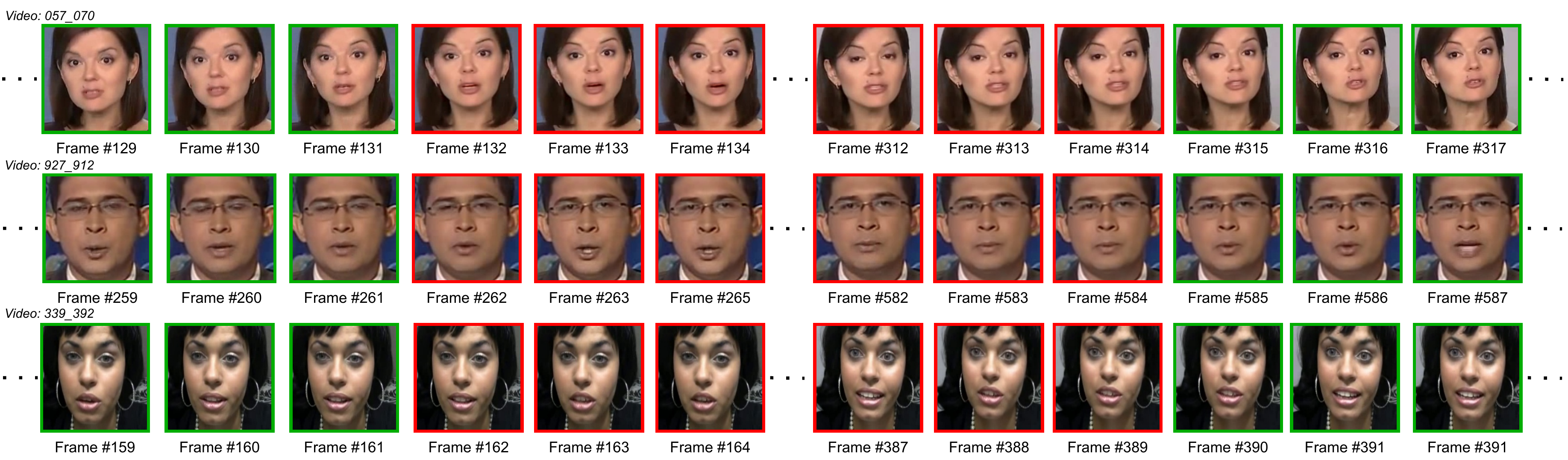}
    \caption{Samples from temporal dataset where segments were carefully selected (hand-crafted) through manual inspection. In this illustration, we focus specifically on where the transitions takes place. This alteration from \textcolor{darkgreen}{\textbf{real}} to \textcolor{red}{\textbf{fake}} frames and vice versa are subtle. The videos start with a real sequence and subtly changes to a fake sequence before going back to real sequence, and they can easily deceive human inspection.}
  \label{fig:temporal_samples}
\end{figure*}

\subsection{Temporal Deepfake Segment Benchmark}\label{sec:benchmark_dataset}
Nearly all deepfake related studies have assumed that all the frames in deepfake videos are from one class, i.e. either fake or real. However, it is possible to have both types of frames within one video. With sophisticated deepfake generation methods such as Neural Textures \cite{neuraltextures}, Deepfacelab \cite{deepfacelab} etc., it is possible to create fake segments that blend masterfully with the neighboring real segments within a video. This makes it very hard even for experienced human eyes to detect and to separate the fake segments from the real ones. Hence, it is important to explore automated temporal segmentation of deepfake videos. 
\vspace{-5mm}
\subsubsection{Dataset}\label{sec:dataset}
The existing deepfake datasets contain videos where all the frames in a video are from one class: `real' or `fake'. To the best of our knowledge, there are currently no publicly available datasets crafted specially for the temporal segmentation task. Therefore, we created the first benchmark dataset with videos that contain fake and real frames to test temporal segmentation. The dataset was created from a resampled subset from the original FaceForensics++ (FF++) dataset \cite{ffpp}. There are five face manipulation techniques within FF++ which we refer to as sub-datasets in our paper. Each sub-dataset has $1,000$ videos in total. These five sub-datasets are prepared based on five different deepfake generation methods: Deepfakes (DF) \cite{deepfakes}, FaceShifter (FSh) \cite{faceshifter}, Face2Face (F2F) \cite{face2face}, NeuralTextures (NT) \cite{neuraltexutres}, FaceSwap (FS) \cite{faceswap}. Since the distribution of any copy of the original dataset is limited, we will publish the code and necessary files to regenerate the temporal dataset instead of publishing the videos. 

Our dataset comes in two main parts: 1) videos with hand-crafted fake segments and 2) videos with randomly chosen fake segments. 

\textbf{Hand-crafted fake-segments}: Since transitioning from real to fake frames and vice versa can cause heavy temporal artifacts, it makes the videos unrealistic if the transition is not subtle. Only videos of \textit{Neural Textures} and \textit{Face2Face} were used to create this part of the dataset, as these methods provide the opportunity for a seamless transition. We manually selected the fake segments to make sure that the changes in sequences are most realistic. Some samples of the NT videos are visualized in Figure \ref{fig:temporal_samples}. The subtlety of the alteration even fools the careful human eyes. There are $100$ videos in each sub-dataset (NT and F2F) where each video contains one fake segment.

\textbf{Randomly chosen fake-segments}: This part contains all the five subdatasets from FF++. Each sub-dataset contains randomly selected $100$ videos. We have $500$ videos in total with one fake segment, and another $500$ videos with two fake segments. 
For videos with one fake segment, we have selected a random starting point in the first half of the video and a random choice from $125$, $150$, and $175$ frames for the length of the fake segment. A similar strategy is also selected for videos with two fake segments. Here, the first fake segment starts at a random position within the first $125$ frames and the second fake segment starts at a random position within the first $75$ frames in the second half of the video. The lengths of the fake segments here are randomly chosen. 

 On average, $24.3\%$ of frames were fake in the videos with one fake segment and the ratio is $41.1\%$ for videos with two fake segments. Average length (number of frames) of a video is $633.9$. Detailed break-down of these ratio and the length of the videos for each deepfake generation method in the dataset are reported in Table \ref{tab:benchmarkstats}.
 
\subsubsection{Evaluation: Intersection over Union (IoU)}\label{sec:ioubaseline}

Intersection over Union (IoU) is proposed to evaluate the temporal segmentation map. This metric is most commonly used to evaluate the fit of object detection bounding boxes \cite{rezatofighi2019generalized, rahman2016optimizing}. 1-D variations of IoU has been adopted for time series segment analysis, which we will be utilizing. 

Let the ground truth map be $GT_{map} = \{ R R R R R R F F F R R ... \}$ and predicted segmentation map be $P_{map} = \{ R R R R R R F F F R R ... \}$. Both are 1-D vectors of equal length with a predicted Boolean class ($R$ or $F$) for each frame in the video.
\begin{equation}
    IoU = \frac{Intersection}{Union}
    = \frac{|GT_{map} \cap P_{map}|}{| GT_{map} \cup P_{map} |}
\end{equation}

IoU falls in the range $[0, 1]$; where the greater the value, the better the predicted segment map. Although the theoretical lower bound of IoU is zero, in practice it is useful to understand how a random guessing algorithm will be scored. For a random guessing algorithm with probability $p=0.5$ for each class in a binary classification problem, we have $IoU = 1/3$. This will be the random guessing baseline for IoU in our context.

\section{Results}

\begin{table}[]
\centering
\begin{tabular}{lllll}
    \hline
    \multicolumn{1}{c|}{Model } & \multicolumn{2}{c|}{\textbf{F2F}} & \multicolumn{2}{c}{\textbf{NT}} \\ \cline{2-5} 
    \multicolumn{1}{c|}{trained on} & IoU & \multicolumn{1}{l|}{AUC} & IoU & AUC \\ \hline \hline
    \multicolumn{1}{c|}{Deepfakes (DF, {\footnotesize ours})} & 0.948 & \multicolumn{1}{l|}{0.964} & 0.684 & 0.744 \\
    \multicolumn{1}{c|}{Face Shifter (FSh, {\footnotesize ours})} & 0.943 & \multicolumn{1}{l|}{0.962} & 0.637 & 0.699 \\
    \multicolumn{1}{c|}{Face2Face (F2F, {\footnotesize ours})} & \textbf{0.980} & \multicolumn{1}{l|}{\textit{0.987}} & 0.738 & 0.794 \\
    \multicolumn{1}{c|}{Neural Textures (NT, {\footnotesize ours})} & 0.943 & \multicolumn{1}{l|}{0.970} & \textbf{0.931} & \textbf{0.960} \\
    \multicolumn{1}{c|}{Face Swap (FS, {\footnotesize ours})} & 0.954 & \multicolumn{1}{l|}{0.969} & 0.553 & 0.607 \\ \hline
    \multicolumn{1}{c|}{\makecell{FF++ (CADDM \cite{caddm})}} & 0.942 & \multicolumn{1}{l|}{\textbf{0.989}} & 0.756 & 0.943 \\ \hline
    \multicolumn{1}{c|}{\textbf{FF++ (ours)}} & \textit{0.970} & \multicolumn{1}{l|}{0.984} & \textit{0.930} & \textit{0.953} \\ \hline
    
\end{tabular}
\caption{Results for temporal segmentation on the proposed temporal dataset with \textbf{hand-crafted fake-segments}. Each row indicates results from a model trained on a specific training sub-dataset; we have trained models with FaceForensics++ (FF++) and the five sub-datasets within FF++. We compare our results with CADDM\cite{caddm}, as shown in the second last row. The last row presents the results for the model trained on the full FF++ dataset. The columns represent the data we have tested our models on; we have tested the models on the two sub-datasets from the hand-crafted temporal segments: NT and F2F. We report IoU and AUC metrics, where the best value in a column is represented in \textbf{bold}, and the second-best value is represented in \textit{italic}.}
\label{tab:manual_results}
\end{table}

\setlength{\tabcolsep}{2pt}  
\renewcommand{\arraystretch}{1.5}
\begin{table*}
    \centering
    \scriptsize
    \begin{tabular}{c|rrrr|rrrr|rrrr|rrrr|rrrr|rrrr}
    \hline
      Model
     &
      \multicolumn{4}{c|}{\textbf{DF}} &
      \multicolumn{4}{c|}{\textbf{FSh}} &
      \multicolumn{4}{c|}{\textbf{F2F}} &
      \multicolumn{4}{c|}{\textbf{NT}} &
      \multicolumn{4}{c|}{\textbf{FS}} &
      \multicolumn{4}{c}{\textbf{FF++ (Average)}} \\ \cline{2-25} 
       trained
     &
      \multicolumn{2}{c|}{One seg} &
      \multicolumn{2}{c|}{Two seg} &
      \multicolumn{2}{c|}{One seg} &
      \multicolumn{2}{c|}{Two seg} &
      \multicolumn{2}{c|}{One seg} &
      \multicolumn{2}{c|}{Two seg} &
      \multicolumn{2}{c|}{One seg} &
      \multicolumn{2}{c|}{Two seg} &
      \multicolumn{2}{c|}{One seg} &
      \multicolumn{2}{c|}{Two seg} &
      \multicolumn{2}{c|}{One seg} &
      \multicolumn{2}{c}{Two seg} \\ \cline{2-25} 
       on
     &
      \multicolumn{1}{l}{IoU} &
      \multicolumn{1}{l|}{AUC} &
      \multicolumn{1}{l}{IoU} &
      \multicolumn{1}{l|}{AUC} &
      \multicolumn{1}{l}{IoU} &
      \multicolumn{1}{l|}{AUC} &
      \multicolumn{1}{l}{IoU} &
      \multicolumn{1}{l|}{AUC} &
      \multicolumn{1}{l}{IoU} &
      \multicolumn{1}{l|}{AUC} &
      \multicolumn{1}{l}{IoU} &
      \multicolumn{1}{l|}{AUC} &
      \multicolumn{1}{l}{IoU} &
      \multicolumn{1}{l|}{AUC} &
      \multicolumn{1}{l}{IoU} &
      \multicolumn{1}{l|}{AUC} &
      \multicolumn{1}{l}{IoU} &
      \multicolumn{1}{l|}{AUC} &
      \multicolumn{1}{l}{IoU} &
      \multicolumn{1}{l|}{AUC} &
      \multicolumn{1}{l}{IoU} &
      \multicolumn{1}{l|}{AUC} &
      \multicolumn{1}{l}{IoU} &
      \multicolumn{1}{l}{AUC} \\ \hline \hline
     \makecell{\textbf{DF} \tiny (ours)} &
      \textbf{0.987} &
      \multicolumn{1}{r|}{0.986} &
      \textbf{0.975} &
      \textit{0.984} &
      0.926 &
      \multicolumn{1}{r|}{0.917} &
      0.886 &
      0.923 &
      0.963 &
      \multicolumn{1}{r|}{0.96} &
      0.937 &
      0.959 &
      0.748 &
      \multicolumn{1}{r|}{0.729} &
      0.603 &
      0.723 &
      0.954 &
      \multicolumn{1}{r|}{0.956} &
      0.920 &
      0.954 &
      0.915 &
      \multicolumn{1}{r|}{0.912} &
      0.860 &
      0.910 \\
     \makecell{\textbf{FSh} \tiny (ours)} &
      0.957 &
      \multicolumn{1}{r|}{0.963} &
      0.933 &
      0.958 &
      \textit{0.972} &
      \multicolumn{1}{r|}{\textit{0.984}} &
      \textit{0.961} &
      \textit{0.980} &
      0.971 &
      \multicolumn{1}{r|}{0.969} &
      0.947 &
      0.966 &
      0.764 &
      \multicolumn{1}{r|}{0.748} &
      0.628 &
      0.745 &
      0.966 &
      \multicolumn{1}{r|}{0.968} &
      0.938 &
      0.965 &
      0.926 &
      \multicolumn{1}{r|}{0.928} &
      0.878 &
      0.924 \\
    \makecell{\textbf{F2F} \tiny (ours)} &
      0.970 &
      \multicolumn{1}{r|}{0.976} &
      0.959 &
      0.976 &
      0.971 &
      \multicolumn{1}{r|}{0.978} &
      0.954 &
      0.972 &
      \textbf{0.982} &
      \multicolumn{1}{r|}{\textit{0.986}} &
      \textbf{0.974} &
      \textbf{0.984} &
      0.840 &
      \multicolumn{1}{r|}{0.836} &
      0.739 &
      0.832 &
      \textbf{0.983} &
      \multicolumn{1}{r|}{0.985} &
      0.966 &
      0.981 &
      0.950 &
      \multicolumn{1}{r|}{0.953} &
      0.918 &
      0.950 \\
    \makecell{\textbf{NT} \tiny (ours)} &
      0.971 &
      \multicolumn{1}{r|}{0.985} &
      0.961 &
      0.980 &
      0.969 &
      \multicolumn{1}{r|}{0.983} &
      0.958 &
      0.979 &
      0.963 &
      \multicolumn{1}{r|}{0.980} &
      0.955 &
      0.977 &
      \textit{0.949} &
      \multicolumn{1}{r|}{\textbf{0.974}} &
      \textit{0.933} &
      \textit{0.966} &
      0.946 &
      \multicolumn{1}{r|}{0.970} &
      0.932 &
      0.965 &
      \textit{0.960} &
      \multicolumn{1}{r|}{\textit{0.979}} &
      \textit{0.949} &
      \textit{0.974} \\
    \makecell{\textbf{FS} \tiny (ours)} &
      0.941 &
      \multicolumn{1}{r|}{0.935} &
      0.898 &
      0.932 &
      0.963 &
      \multicolumn{1}{r|}{0.962} &
      0.931 &
      0.955 &
      0.970 &
      \multicolumn{1}{r|}{0.970} &
      0.950 &
      0.969 &
      0.679 &
      \multicolumn{1}{r|}{0.679} &
      0.514 &
      0.642 &
      \textit{0.981} &
      \multicolumn{1}{r|}{\textit{0.987}} &
      \textbf{0.967} &
      \textbf{0.983} &
      0.904 &
      \multicolumn{1}{r|}{0.901} &
      0.842 &
      0.898 \\ \hline
    \makecell{\textbf{FF++} \\ \textbf{\tiny (CADDM)}} &
      0.966 &
      \multicolumn{1}{r|}{\textit{0.987}} &
      0.957 &
      \textbf{0.987} &
      0.943 &
      \multicolumn{1}{r|}{0.981} &
      0.903 &
      0.977 &
      0.935 &
      \multicolumn{1}{r|}{0.971} &
      0.891 &
      0.971 &
      0.745 &
      \multicolumn{1}{r|}{0.883} &
      0.598 &
      0.874 &
      0.974 &
      \multicolumn{1}{r|}{0.986} &
      \textit{0.970} &
      \textit{0.986} &
      0.913 &
      \multicolumn{1}{r|}{0.962} &
      0.864 &
      0.869 \\ \hline
    \makecell{\textbf{FF++} \\ (\textbf{ours})} &
      \textit{0.974} &
      \multicolumn{1}{r|}{\textbf{0.988}} &
      \textit{0.962} &
      0.982 &
      \textbf{0.974} &
      \multicolumn{1}{r|}{\textbf{0.989}} &
      \textbf{0.962} &
      \textbf{0.982} &
      \textit{0.975} &
      \multicolumn{1}{r|}{\textbf{0.988}} &
      \textit{0.965} &
      \textit{0.983} &
      \textbf{0.959} &
      \multicolumn{1}{r|}{\textit{0.972}} &
      \textbf{0.938} &
      \textbf{0.967} &
      0.975 &
      \multicolumn{1}{r|}{\textbf{0.988}} &
      0.955 &
      0.978 &
      \textbf{0.971} &
      \multicolumn{1}{r|}{\textbf{0.985}} &
      \textbf{0.957} &
      \textbf{0.979} \\ \hline
    \end{tabular}
    \caption{Results for temporal segmentation on the proposed benchmark temporal deepfake dataset with \textbf{randomly chosen fake-segments}. Structure of this table is similar to of Table \ref{tab:manual_results}. For each test sub-dataset we have tested separately for videos with one fake-segment and two fake-segments from our proposed benchmark dataset (see section \ref{sec:benchmark_dataset}). }
    \label{tab:temporal-result}
\end{table*}
\renewcommand{\arraystretch}{1}

\subsection{Experimental settings}
\textbf{Dataset.} \quad
Based on the recent deepfake detection methods we have used FaceForensics++ \cite{ffpp} (FF++, see Section \ref{sec:dataset}) for training our models. Among the $1,000$ in each sub-dataset within FF++, we have used 800 for training and validation, and we tested on the remaining 200. We have experimented with the videos with compression level `c23' and all the frames were used in the training and testing, i.e., no frames were skipped.

To evaluate the temporal segmentation performance, we have used our proposed novel benchmark dataset (see Section \ref{sec:benchmark_dataset}) for testing the temporal segmentation performance. This dataset is generated based on the original FF++ dataset, where each video contains both real and fake segments. There are 5 sub-datasets within the temporal dataset, the same as the original FF++, each having $100$ videos with an average of $633.9$ frames per video, $24.3\%$ fake frames for videos with one fake segment, and $41.1\%$ fake frames for videos with two fake segments. We also experimented with other popular datasets: FF++, CelebDF\cite{celebdf}, DFDC\cite{dfdc} and WildDeepFakes\cite{wilddeepfakes}. These datasets were used to compare our method's performance in terms of traditional deepfake detection experiments: classical (same-dataset) deepfake detection and generalizability (cross-dataset). 

For face extraction and alignment, we have used DLIB\cite{dlib}, and aligned face images were resized to $224\times224$ for all frames in the train and test sets. \\

\textbf{Settings. Vision Transformer (ViT) and Shift and Scaling (SSF)}. \quad
A simple set of preprocessing steps including extraction of the frames and cropping the face region is done as data preparation for ViT. We use the ViT-B/16\cite{vitb16} architecture as the backbone for our ViT model and trained it using four A100 GPUs with a batch size of $64$ for $80$ epochs. The optimization algorithm used was `AdamW' with a weight decay of $0.05$. A warm-up strategy was applied for the learning rate, starting with $1e^{-7}$ for the first five epochs and then increasing linearly to $1e^{-3}$. The dropout probability was set to $0.1$, and the input image size was set to $224\times224$. To improve performance, an exponential moving average was used with a decay rate of $0.99992$ together with automatic mixed precision. The model was initialized with pre-trained weights on Imagenet-21K-SSF. \\

\textbf{Timeseries settings.} \quad
The dimension of the feature vector for each frame from the ViT was $768$. After widowing these vectors as shown in Figure \ref{fig:archi}  the input dimension for the Timeseries Transformer (TsT) was $(W, 768)$. In our experiments, we used $W=5$ however it is also possible to use different values for $W$. There are a total of $8$ transformer blocks in the TsT, with an $8$-headed attention connection. Each attention-head's dimension is $512$. After the transformer blocks, we have a one-dimensional global average pool prior to the MLP head. We have used batch size of $64$, `categorical cross-entropy' as the loss and `Adam' as the optimizer with $1e^{-4}$ learning rate for training the TsT. We use an early stopping technique with patience of $10$ to speed up the training procedure. 

\subsection{Temporal segmentation analysis}
We have used our proposed benchmark dataset to test our method for the temporal segmentation problem, where we try to classify deepfake videos at the frame-level instead of video-level. The metrics we use to measure the performance for temporal segmentation are Intersection over Union (IoU) and Area under the ROC Curve (AUC). The baseline IoU (for random guessing of the class of a frame) is $1/3$ as shown in Section \ref{sec:ioubaseline}. We have trained six separate models on six training sets: FaceForensics++ (FF++) and the five sub-datasets within FF++ i.e. Deepfakes (DF), Face-Shifter (FSh), Face2Face (F2F), Neural Textures (NT) and FaceSwap (FS). Similarly, we report the results for each model on the six test sets (FF++ and its five sub-datasets) in Tables \ref{tab:manual_results} and \ref{tab:temporal-result}. We compare our results with the CADDM \cite{caddm} where we have used the original implementation and model weights published by the authors with a slight modification to compute frame-level AUC and IoU values.

As seen, each model does very well when it was tested on the test-set from the same dataset as it was trained on, hence the results on the diagonals are either the best or the second-best in every column while the second-best results are only lower in the range from $0.001$ to $0.009$. As expected, the model trained on the whole FF++ is the overall best-performing model. In Table \ref{tab:manual_results} and \ref{tab:temporal-result} we can see that our detection method outperforms the latest state-of-the-art, CADDM \cite{caddm} method in the temporal segmentation task.

However, the results of the other five models give us some important findings. Of the five sub-datasets in FF++, three were made with a face swapping technique (DF, FSh and FS) and the other two were made with face reenactment (F2F and NT). We can see that the models trained in F2F and NT perform better than the other three models. Since face reenactment deepfakes are devoid of strong artifacts compared to face swapping deepfakes, models trained on face reenactment methods tend to generalize well to other methods. Similarly, models trained on the face swapping methods do not generalize well to face re-enactment test data as seen on the `F2F' and `NT' columns in both Table \ref{tab:manual_results} and \ref{tab:temporal-result}. Overall, out of the five subdatasets of FF++, NT is the best one to train a detection model if others are unavailable.

Also, our method does significantly better with these two sub-datasets compared to CADDM \cite{caddm} which indicates the necessity to learn temporal features to predict the transition between real and fake segments more effectively.
\renewcommand{\arraystretch}{1.3}
\begin{table}
    \centering
    \small
    \begin{tabular}{c|lllll|llll}
    \hline
     & \textbf{DF} & \textbf{FSh} & \textbf{F2F} & \textbf{NT} & \textbf{FS} & \textbf{FF++} & \textbf{C-DF} & \textbf{DFDC} & \textbf{WDF} \\ \hline \hline
    \textbf{DF} & \textbf{0.993} & 0.965 & 0.975 & 0.830 & 0.967 & 0.946 & 0.590 & 0.558 & 0.631 \\
    \textbf{FSh} & 0.980 & \textit{0.990} & 0.978 & 0.848 & 0.980 & 0.955 & 0.609 & 0.535 & 0.619 \\
    \textbf{F2F} & \textit{0.990} & \textbf{0.993} & \textbf{0.990} & 0.917 & \textit{0.993} & \textit{0.977} & 0.670 & 0.603 & 0.676 \\
    \textbf{NT} & 0.973 & 0.967 & 0.967 & \textit{0.965} & 0.960 & 0.967 & \textit{0.715} & \textit{0.626} & \textit{0.693} \\
    \textbf{FS} & 0.960 & 0.975 & 0.982 & 0.770 & \textbf{0.995} & 0.936 & 0.584 & 0.512 & 0.611 \\ \hline
    \textbf{FF++} & 0.985 & 0.983 & \textit{0.983} & \textbf{0.967} & 0.983 & \textbf{0.982} & \textbf{0.790} & \textbf{0.667} & \textbf{0.703} \\ \hline
    \end{tabular}
    \caption{Results (in AUC) for video level classification. Similar to Tables \ref{tab:manual_results} and \ref{tab:temporal-result}, each row represents models trained on specific training data and the columns constitute the test data. Along with FF++ and its sub-datasets we have tested each model on other datasets such as CelebDF (C-DF), DFDC, and WildDeepFakes (WDF).}
    \label{tab:video-level}
\end{table}
\renewcommand{\arraystretch}{1}

\setlength{\tabcolsep}{1em}
\renewcommand{\arraystretch}{1.2}
\begin{table}
    \centering
    \small
    \begin{tabular}{cccc}
        \hline
        \textbf{Method} & \textbf{CelebDF} & \textbf{FF++} & \textbf{NT} \\ \hline \hline
        \textbf{Xception}\cite{xception} & 0.653 & 0.997 & 0.842 \\
        \textbf{SRM}\cite{srm} & 0.659 & 0.969 & \textit{0.943} \\
        \textbf{SPSL}\cite{spsl} & 0.724 & 0.969 & 0.805 \\
        \textbf{MADD}\cite{madd} & 0.674 & \textbf{0.998} & -- \\
        \textbf{SLADD}\cite{sladd} & 0.797 & \textit{0.984} & -- \\
        \textbf{CADDM}\cite{caddm} & \textbf{0.931} & 0.998 & 0.837 \\ \hline
        \textbf{Ours (ViT+TsT)} & 0.790 & 0.982 & \textbf{0.967} \\ \hline
    \end{tabular}
    \caption{Comparison with other state-of-the-art methods in terms of video-level AUC. Models were trained on FF++ and evaluated on CelebDF, FF++ and NT. Results for the other methods were taken from their own paper or github repository page. NT is the most challenging deepfake generation technique in terms of both temporal segmentation and video level detection. Our method significantly outperforms recent methods in detecting NT fakes while performing competitively on other datasets.}
    \label{tab:comparison}
\end{table}
\renewcommand{\arraystretch}{1}

\subsection{Video-level classification and Generalizability}
The classical approach to deepfake detection has always been to predict the class (real or fake) of a deepfake video i.e. to make video-level predictions. We performed experiments on the test data from the original datasets and reported the results in Table \ref{tab:video-level} using AUC as the metric. We also measure our models' performance on test data from datasets outside of FF++: CelebDF (C-DF), DFDC, and WildDeepFakes (WDF). Results for the sub-datasets of FF++ (DF, FSh, F2F, NT and FS) follow the results of the temporal analysis where the diagonal values are the best or the second-best in a column, i.e. models when tested on data from the same sub-dataset generally perform very well. And, similar to the previous results (i.e. temporal segmentation), we see that models trained on face re-enactment data (F2F and NT) perform better than other sub-dataset-models when tested on unseen data, i.e. these two models generalize well in comparison with other models. However, we see the best results from tests on CelebDF, DFDC, and WildDeepFakes from the model trained on the full FF++ dataset. 

It is also noticeable that the AUC scores for DFDC and WildDeepFakes are lower compared to CelebDF. While CelebDF is a dataset with videos made solely by the face swapping technique, DFDC is a combination of multiple methods such as face swapping (Deepfake Autoencoder and Morphable-mask), Neural talking-heads and GAN-based methods. WildDeepFakes dataset contains videos from the Internet which may contain videos generated using a variety of methods. Some of these methods are totally unseen due to their absence in the FF++ dataset.

\setlength{\tabcolsep}{0.3em}
\renewcommand{\arraystretch}{1.2}
\begin{table}[]
    \centering
    \scriptsize
    \begin{tabular}{c|cccccccccc}
    \hline
     \makecell{Length \\ (seconds)} & \textbf{1.0}     & \textbf{2.0}     & \textbf{3.0}     & \textbf{4.0}     & \textbf{5.0}     & \textbf{6.0}     & \textbf{7.0}     & \textbf{8.0}     & \textbf{9.0}     & \textbf{10.0}    \\ \hline
    \textbf{IoU}  & 0.961 & 0.962 & 0.962 & 0.963 & 0.963 & 0.965 & 0.965 & 0.967 & 0.967 & 0.969 \\
    \textbf{AUC}  & 0.963 & 0.976 & 0.979 & 0.982 & 0.982 & 0.984 & 0.984 & 0.985 & 0.984 & 0.985 \\ \hline
    \end{tabular}
\caption{IoU and AUC of the our proposed method across different lengths of deepfake segments. Our approach is largely robust to variation in the length of injected deepfake segment.}
\label{tab:phoneme}
\end{table}
\renewcommand{\arraystretch}{1}

We further evaluate and compare our results with the latest state-of-the-art methods in Table \ref{tab:comparison} using the model trained on FF++ and tested on FF++, CelebDF, and NT. Our detection method outperforms the state-of-the-art methods in detecting videos generated by the NT method which is the most difficult method in FF++. Our method also performs very competitively with the most recent state-of-the-art deepfake detection methods \cite{caddm, sladd, madd} and outperforms most methods in video-level predictions in both same-dataset and cross-dataset scenarios. Our method comprehensively exceeds the performance of several state-of-the-art methods such as Xception \cite{ffpp}, SRM \cite{srm}, SPSL \cite{spsl}, MADD \cite{madd} and others (not included in Table \ref{tab:comparison}) \cite{f3net, smil, twobranch} in generalizability (i.e. cross-dataset/CelebDF). These results demonstrate that our method can also be used with high confidence for traditional deepfake detection and for unseen data (i.e. video-level) alongside temporal segmentation despite not being optimized for this objective.

\subsection{Varying Lengths of Fake Segments}
Our proposed method is effective in identifying even short segments of deepfake that can significantly alter the message conveyed by a video. To evaluate the performance of our method, we conducted experiments on a test set comprising $100$ videos with varying lengths of fake segments, and the results are presented in Table \ref{tab:phoneme} and Figure \ref{fig:phoneme}. Specifically, we create fake segments with durations ranging from $0.2$ seconds to $19$ seconds, with an increase of $0.2$ seconds, and calculate the average IoU and AUC over the 100 videos. The increment of $0.2$ seconds is the average duration of two phonemes in English \cite{fant1991durational}, which we assume to be the unit duration for a fake segment. In particular, we did not use the smoothing of noisy frames in this experiment.

Our method achieves high accuracy in detecting very short fake-segments with a duration of less than $1.0$ second, yielding an AUC value of over $0.91$. Furthermore, as the length of the fake segment increases, our method performs even better in terms of AUC and IoU. This experiment provides evidence that our proposed method can identify even the slightest alterations in very short fake-segments, highlighting its effectiveness in detecting deepfake videos.

\subsection{Ablation study}

We conducted experiments to evaluate the effectiveness of our proposed method without TsT and the smoothing algorithm for both temporal segmentation detection and video-level detection. The model was trained on the full FF++ training data and tested on the proposed temporal segmentation benchmark dataset and FF++ test set for temporal segmentation and video-level detection, respectively. We used a MLP head on the ViT to classify frames for the experiment where the TsT was not included. Our ablated model achieved great results in both test sets, which are reported in Table \ref{tab:ablation}. To provide a better comparison, we also reported the results from the full model with the TsT and smoothing algorithm. While we observe that the ViT already performs very well, a significant improvement can be seen in both temporal and video-level performance with the inclusion of the TsT and the smoothing algorithm.

Another ablation study is on varying window sizes for the TsT. In TsT we use a sliding window technique to accumulated features of multiple frames within a window so that the TsT can learn temporal features. We have experimented with varying window sizes in terms of number of frames, accompanying with varying overlap values in the sliding window method. Based on the results from these experiments on both frame-level and video-level predictions as reported in Tables \ref{tab:ablation_window_frame} and \ref{tab:ablation_window_video} we have selected the window size of $5$ frames with ovelap of $4$ frames to be the optimal parameters.

\setlength{\tabcolsep}{0.4em}
\renewcommand{\arraystretch}{1.2}
\begin{table}
    \centering
    \scriptsize
    \begin{tabular}{c|ccc|cc}
        \hline
         \multirow{2}{*}{\makecell{Model \\ trained on}} & \multicolumn{3}{c|}{\textbf{Temporal Evaluation (IoU)}} & \multicolumn{2}{c}{\textbf{Video level (AUC)}} \\ \cline{2-6} 
          & ViT & ViT+TsT & ViT+TsT+Smooth. & ViT & ViT+TsT \\ \hline \hline
        \textbf{DF} & 0.960 & 0.967 (\textbf{+0.007}) & 0.974 (\textbf{+0.014}) & 0.973 & 0.985 (\textbf{+0.012}) \\
        \textbf{FSh} & 0.960 & 0.967 (\textbf{+0.007}) & 0.974 (\textbf{+0.014}) & 0.973 & 0.983 (\textbf{+0.010}) \\
        \textbf{F2F} & 0.956 & 0.968 (\textbf{+0.012}) & 0.975 (\textbf{+0.019}) & 0.973 & 0.983 (\textbf{+0.010}) \\
        \textbf{NT} & 0.933 & 0.948 (\textbf{+0.015}) & 0.959 (\textbf{+0.026}) & 0.965 & 0.967 (\textbf{+0.002}) \\
        \textbf{FS} & 0.951 & 0.965 (\textbf{+0.014}) & 0.975 (\textbf{+0.024}) & 0.973 & 0.983 (\textbf{+0.010}) \\ \hline
        \textbf{FF++} & 0.952 & 0.964 (\textbf{+0.012}) & 0.971 (\textbf{+0.019}) & 0.971 & 0.982 (\textbf{+0.011}) \\ \hline
    \end{tabular}

    \caption{Ablation study on temporal segmentation of deepfakes and video-level classification. For temporal evaluation, results (IoU) from three experiments are reported: from Vision Transformer (ViT) only, with Timeseries Transformer (TsT) and also including smoothing. Video level evaluation (in AUC) is reported for ViT, and ViT with TsT. Changes in the results are reported in bold and are in parentheses.}
    \label{tab:ablation}
\end{table}
\renewcommand{\arraystretch}{1}

\section{Discussion}\label{sec:discussion}

While most methods tackle deepfake detection at the video-level, we propose a robust and generalizable method that can produce results at the frame, segment and entire video level. This allows maximal flexibility in analyzing content for the presence of deepfakes and additionally provides comparison points for future research along these related but separate evaluation protocols.

Our method is based on supervised pretraining of the image encoder, which limits computational requirements in two forms. First, the image encoder is trained independently on individual video frames with frame-level supervision, nullifying the need to learn temporal relationships between frames. Second, a large part of the backbone is frozen and initialized using readily available weights from ImageNet, considerably reducing the computational cost of obtaining a deepfake related representation in the encoder. The proposed method achieves robust IoU metrics across the proposed single and multi-segment deepfakes, while maintaining competitive performance on video-level deepfake detection and generalized deepfake detection.

{\small
\bibliographystyle{ieee}
\bibliography{egbib}
}
\clearpage
\appendix

\section{Appendix}

\renewcommand{\arraystretch}{1.3}

\begin{table}[!h]
    \centering
    \begin{tabular}{lcrrr}
    \hline
         & \multicolumn{1}{c}{\multirow{2}{*}{}} & \multicolumn{2}{c}{\textbf{Ratio of fake frames}} & \multicolumn{1}{l}{\textbf{Avg. Length}} \\ \cline{3-4}
         & \multicolumn{1}{c}{} & \multicolumn{1}{l}{\textbf{One seg.}} & \multicolumn{1}{l}{\textbf{Two seg.}} & \multicolumn{1}{l}{} \\ \hline
        \multicolumn{1}{l|}{M} & \textbf{NT, F2F} & 0.363 & N/A & 193.23 \\ \hline
        \multicolumn{1}{l|}{\parbox[t]{2mm}{\multirow{5}{*}{\rotatebox[origin=c]{90}{Random}}}} & \textbf{DF} & 0.231 & 0.389 & 668.5 \\
        \multicolumn{1}{l|}{} & \textbf{FSh} & 0.231 & 0.389 & 668.5 \\
        \multicolumn{1}{l|}{} & \textbf{F2F} & 0.233 & 0.393 & 662.0 \\
        \multicolumn{1}{l|}{} & \textbf{NT} & 0.264 & 0.445 & 585.3 \\
        \multicolumn{1}{l|}{} & \textbf{FS} & 0.264 & 0.445 & 585.3 \\ \hline
        \multicolumn{1}{l}{} & \textbf{Average} & \textbf{0.243} & \textbf{0.411} & \textbf{633.9}
    \end{tabular}
    \caption{Ratio of fake frames and average length of videos in the benchmark dataset. This benchmark dataset is based on FaceForensics++ (FF++) and has the same sub-datasets as FF++. The ratio of fake frames differs among sub-datasets due to the original fake videos having different number of total frames. The average length is calculated in terms of the number of frames in a video. Each segment of fake frames is contiguous.}
    \label{tab:benchmarkstats}
\end{table}

\begin{table*}[ht]
\small
\centering
    \begin{tabular}{l|rr|rr|rr|rr|rr|rr}
    \hline
    ~ & \multicolumn{2}{c|}{\textbf{DF}} & \multicolumn{2}{c|}{\textbf{FSh}} & \multicolumn{2}{c|}{\textbf{F2F}} & \multicolumn{2}{c|}{\textbf{NT}} & \multicolumn{2}{c|}{\textbf{FS}} & \multicolumn{2}{c}{\textbf{FF++}} \\ \cline{2-13} 
    ~ & \multicolumn{1}{c}{One seg} & \multicolumn{1}{c|}{Two seg} & \multicolumn{1}{c}{One seg} & \multicolumn{1}{c|}{Two seg} & \multicolumn{1}{c}{One seg} & \multicolumn{1}{c|}{Two seg} & \multicolumn{1}{c}{One seg} & \multicolumn{1}{c|}{Two seg} & \multicolumn{1}{c}{One seg} & \multicolumn{1}{c|}{Two seg} & \multicolumn{1}{c}{One seg} & \multicolumn{1}{c}{Two seg} \\ \hline \hline
    \textbf{DF} & \textbf{0.993} & \textbf{0.987} & 0.961 & 0.939 & 0.981 & 0.967 & 0.856 & 0.752 & 0.977 & 0.958 & 0.956 & 0.925 \\
    \textbf{FSh} & 0.978 & 0.965 & \textit{0.986} & \textit{0.98} & 0.985 & 0.973 & 0.866 & 0.772 & 0.983 & 0.968 & 0.962 & 0.935 \\
    \textbf{F2F} & 0.985 & 0.979 & 0.986 & 0.977 & \textbf{0.991} & \textbf{0.987} & 0.913 & 0.850 & 0.992 & 0.983 & 0.974 & 0.957 \\
    \textbf{NT} & 0.985 & 0.980 & 0.984 & 0.979 & 0.981 & 0.977 & \textit{0.974} & \textit{0.965} & 0.972 & 0.965 & \textit{0.980} & \textit{0.974} \\
    \textbf{FS} & 0.914 & 0.859 & 0.960 & 0.930 & 0.972 & 0.952 & 0.761 & 0.592 & \textbf{0.993} & \textbf{0.985} & 0.922 & 0.868 \\ \hline
    \textbf{FF++} & \textit{0.987} & \textit{0.981} & \textbf{0.987} & \textbf{0.98} & \textit{0.987} & \textit{0.982} & \textbf{0.979} & \textbf{0.968} & \textit{0.987} & \textit{0.977} & \textbf{0.986} & \textbf{0.978} \\ \hline
    \end{tabular}
\caption{Results in terms of accuracy for temporal segmentation on the proposed benchmark temporal deepfake dataset. This table is supplementary and identical in organization to Table \ref{tab:temporal-result} in the main paper. Each row indicates a model trained on a specific training sub-dataset; we have trained models with FaceForensics++ (FF++) and the five sub-datasets within FF++ i.e. Deepfakes (DF), Face-Shifter (FSh), Face2Face (F2F), Neural Textures (NT) and FaceSwap (FS). We report the best value in a column in \textbf{bold} and the second-best in \textit{italic}.}
\label{tab:supp_temporal_accuracy}
\end{table*}

\begin{table}[!ht]
\scriptsize
\centering
    \begin{tabular}{l|ccccc|cccc}
    \hline
     & \multicolumn{1}{l}{\textbf{DF}} & \multicolumn{1}{l}{\textbf{FSh}} & \multicolumn{1}{l}{\textbf{F2F}} & \multicolumn{1}{l}{\textbf{NT}} & \multicolumn{1}{l|}{\textbf{FS}} & \multicolumn{1}{l}{\textbf{FF++}} & \multicolumn{1}{l}{\textbf{C-DF}} & \multicolumn{1}{l}{\textbf{DFDC}} & \multicolumn{1}{l}{\textbf{WDF}} \\ \hline \hline
    \textbf{DF} & \textbf{0.993} & 0.965 & 0.975 & 0.83 & 0.968 & 0.917 & 0.301 & 0.550 & 0.625 \\
    \textbf{FSh} & 0.980 & \textit{0.990} & 0.978 & 0.848 & 0.980 & 0.935 & 0.402 & 0.555 & 0.613 \\
    \textbf{F2F} & \textit{0.990} & \textbf{0.993} & \textbf{0.990} & 0.917 & \textit{0.993} & 0.968 & 0.535 & \textit{0.589} & 0.672 \\
    \textbf{NT} & 0.973 & 0.968 & 0.968 & \textit{0.965} & 0.960 & \textit{0.978} & \textit{0.593} & 0.584 & \textit{0.694} \\
    \textbf{FS} & 0.855 & 0.945 & 0.970 & 0.590 & \textbf{0.995} & 0.788 & 0.322 & 0.534 & 0.532 \\ \hline
    \textbf{FF++} & 0.985 & 0.983 & \textit{0.983} & \textbf{0.968} & 0.983 & \textbf{0.987} & \textbf{0.799} & \textbf{0.682} & \textbf{0.694} \\ \hline
    \end{tabular}
\caption{Results (in Accuracy) for video level classification. This table is supplementary and identical in organization to Table \ref{tab:video-level} in the main paper. The columns constitute the test data. Along with FF++ and the sub-datasets of FF++ we have tested each model on other datasets such as CelebDF (C-DF), DFDC, and WildDeepFakes (WDF). The best value in a column is in \textbf{bold} and the second-best is in \textit{italic}.}
\label{tab:supp_video_level_accuracy}
\end{table}

\begin{table}[!ht]
\scriptsize
\centering
\begin{tabular}{l|ccc|cc}
\hline
 & \multicolumn{3}{c|}{\textbf{Temporal Evaluation (Accuracy)}} & \multicolumn{2}{c}{\textbf{Video level (Accuracy)}} \\ \cline{2-6} 
 & ViT & ViT+TsT & ViT+TsT+Algo 1 & ViT & ViT+TsT \\ \hline \hline
\textbf{DF} & 0.981 & 0.983 (\textbf{+0.002}) & 0.987 (\textbf{+0.006}) & 0.973 & 0.985 (\textbf{+0.012}) \\
\textbf{FSh} & 0.981 & 0.983 (\textbf{+0.002}) & 0.987 (\textbf{+0.006}) & 0.973 & 0.983 (\textbf{+0.010}) \\
\textbf{F2F} & 0.982 & 0.984 (\textbf{+0.002}) & 0.987 (\textbf{+0.005}) & 0.973 & 0.983 (\textbf{+0.010}) \\
\textbf{NT} & 0.970 & 0.973 (\textbf{+0.003}) & 0.979 (\textbf{+0.009}) & 0.965 & 0.968 (\textbf{+0.003}) \\
\textbf{FS} & 0.979 & 0.982 (\textbf{+0.003}) & 0.987 (\textbf{+0.008}) & 0.973 & 0.983 (\textbf{+0.010}) \\ \hline
\textbf{FF++} & 0.979 & 0.981 (\textbf{+0.002}) & 0.986 (\textbf{+0.007}) & 0.985 & 0.987 (\textbf{+0.002}) \\ \hline
\end{tabular}
\caption{Results in Accuracy for Ablation study on temporal segmentation of deepfakes and video-level classification. This table is supplementary and identical in organization to Table \ref{tab:ablation} in the main paper. Changes in the results are reported bold and are in brackets.}
\label{tab:ablation_supp}
\end{table}

\begin{table}[!h]
\centering
    \begin{tabular}{ccccccc}
     & \multicolumn{6}{c}{Window Size} \\ \cline{2-7} 
     & \multicolumn{2}{c|}{\textbf{5}} & \multicolumn{2}{c|}{\textbf{10}} & \multicolumn{2}{c}{\textbf{15}} \\ \cline{2-7} 
    Overlap & IoU & \multicolumn{1}{c|}{AUC} & IoU & \multicolumn{1}{c|}{AUC} & IoU & AUC \\ \hline
    \multicolumn{1}{c|}{\textbf{4}} & \textbf{0.974} & \multicolumn{1}{c|}{\textbf{0.988}} & 0.956 & \multicolumn{1}{c|}{0.973} & 0.797 & 0.846 \\
    \multicolumn{1}{c|}{\textbf{3}} & 0.953 & \multicolumn{1}{c|}{0.977} & 0.946 & \multicolumn{1}{c|}{0.975} & 0.806 & 0.848 \\
    \multicolumn{1}{c|}{\textbf{2}} & 0.950 & \multicolumn{1}{c|}{0.976} & 0.953 & \multicolumn{1}{c|}{0.976} & 0.745 & 0.766 \\
    \multicolumn{1}{c|}{\textbf{1}} & \textit{0.971} & \multicolumn{1}{c|}{\textit{0.985}} & 0.958 & \multicolumn{1}{c|}{0.976} & 0.797 & 0.838 \\
    \multicolumn{1}{c|}{\textbf{0}} & 0.958 & \multicolumn{1}{c|}{0.983} & 0.947 & \multicolumn{1}{c|}{0.975} & 0.766 & 0.84 \\ \hline
    \end{tabular}
\caption{Ablation study on varying Window sizes in terms of number of frames in a window and overlap in sliding-window. The values are from frame-level prediction on our proposed temporal segmentation dataset with one fake-segment to solve the temporal segmentation problem. We can notice that a window size of $5$ with overlap of $4$ gives us the optimal results for temporal segmentation.}
\label{tab:ablation_window_frame}
\end{table}

\begin{table}[!h]
\centering
\begin{tabular}{crrrrrr}
 & \multicolumn{6}{c}{Window Size} \\ \cline{2-7} 
 & \multicolumn{2}{c|}{\textbf{5}} & \multicolumn{2}{c|}{\textbf{10}} & \multicolumn{2}{c}{\textbf{15}} \\ \cline{2-7} 
Overlap & \multicolumn{1}{c}{Acc} & \multicolumn{1}{c|}{AUC} & \multicolumn{1}{c}{Acc} & \multicolumn{1}{c|}{AUC} & \multicolumn{1}{c}{Acc} & \multicolumn{1}{c}{AUC} \\ \hline
\multicolumn{1}{c|}{\textbf{4}} & \textit{0.987} & \multicolumn{1}{r|}{\textit{0.982}} & \textbf{0.992} & \multicolumn{1}{r|}{\textbf{0.985}} & 0.982 & 0.959 \\
\multicolumn{1}{c|}{\textbf{3}} & 0.987 & \multicolumn{1}{r|}{0.974} & 0.983 & \multicolumn{1}{r|}{0.972} & 0.983 & 0.966 \\
\multicolumn{1}{c|}{\textbf{2}} & 0.988 & \multicolumn{1}{r|}{0.975} & 0.984 & \multicolumn{1}{r|}{0.977} & 0.980 & 0.976 \\
\multicolumn{1}{c|}{\textbf{1}} & 0.982 & \multicolumn{1}{r|}{0.981} & 0.983 & \multicolumn{1}{r|}{0.974} & 0.985 & 0.969 \\
\multicolumn{1}{c|}{\textbf{0}} & 0.990 & \multicolumn{1}{r|}{0.978} & 0.983 & \multicolumn{1}{r|}{0.974} & 0.980 & 0.944 \\ \hline
\end{tabular}
\caption{Ablation study on varying Window sizes in terms of number of frames in a window and overlap in sliding-window. The values are from video-level prediction. We can notice that a window size of $5$ with overlap of $4$ gives us the second-best results where the results for window size $10$ with overlap of $4$ frames are the best. However, our main goal is to achieve best results in frame-level performance. Hence, we chose the prior parameters for the experiments.}
\label{tab:ablation_window_video}
\end{table}

\renewcommand{\arraystretch}{0.8}

\begin{figure}[!ht]
    \centering
    \includegraphics[width=0.4\textwidth]{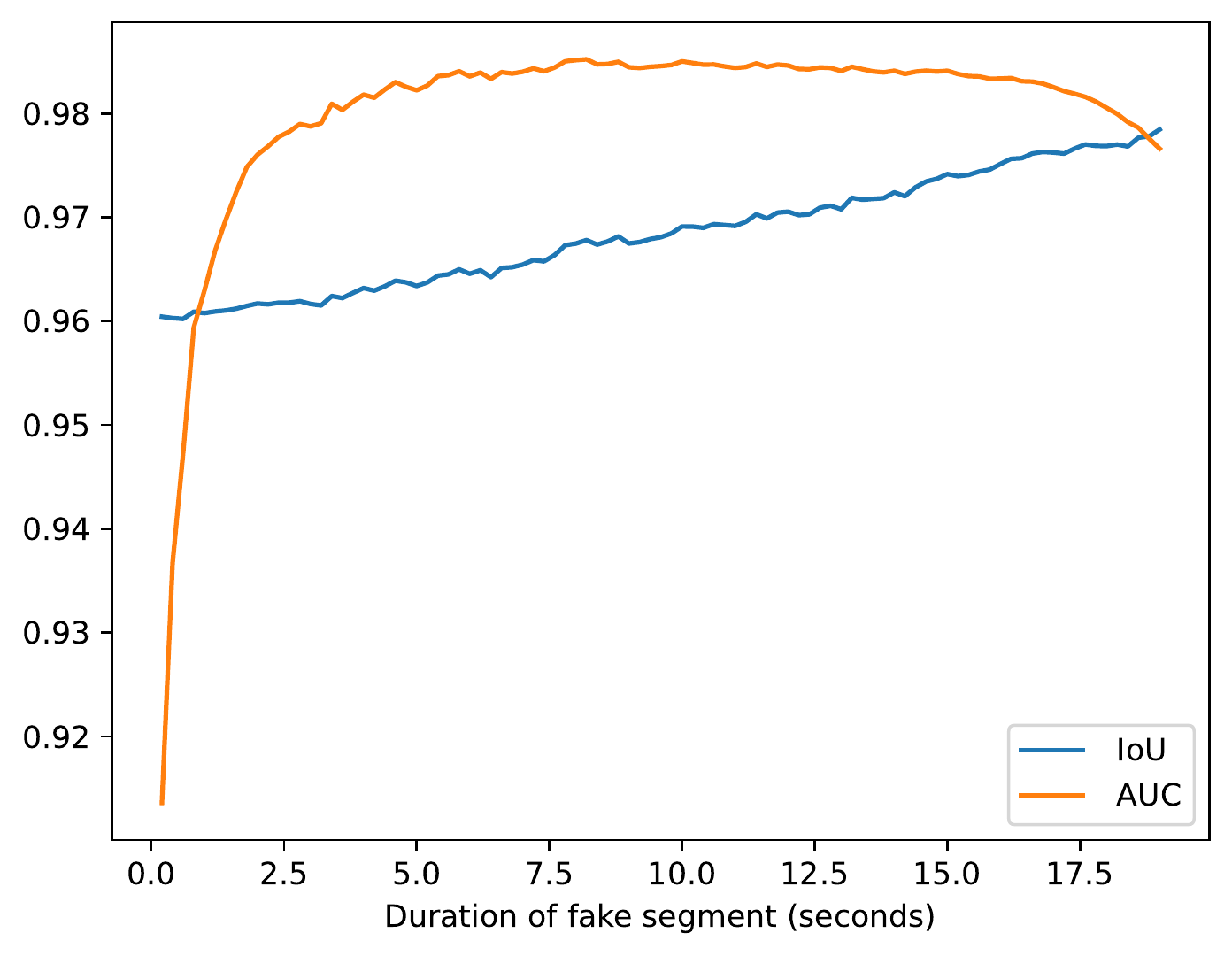}
    \caption{Performance (IoU and AUC) of the proposed approach across different lengths of deepfake segments. This is a visualization of Table \ref{tab:phoneme} with more dense data points.}
    \label{fig:phoneme}
\end{figure}

\begin{figure}[!ht]
    \centering
    \includegraphics[width=0.49\textwidth]{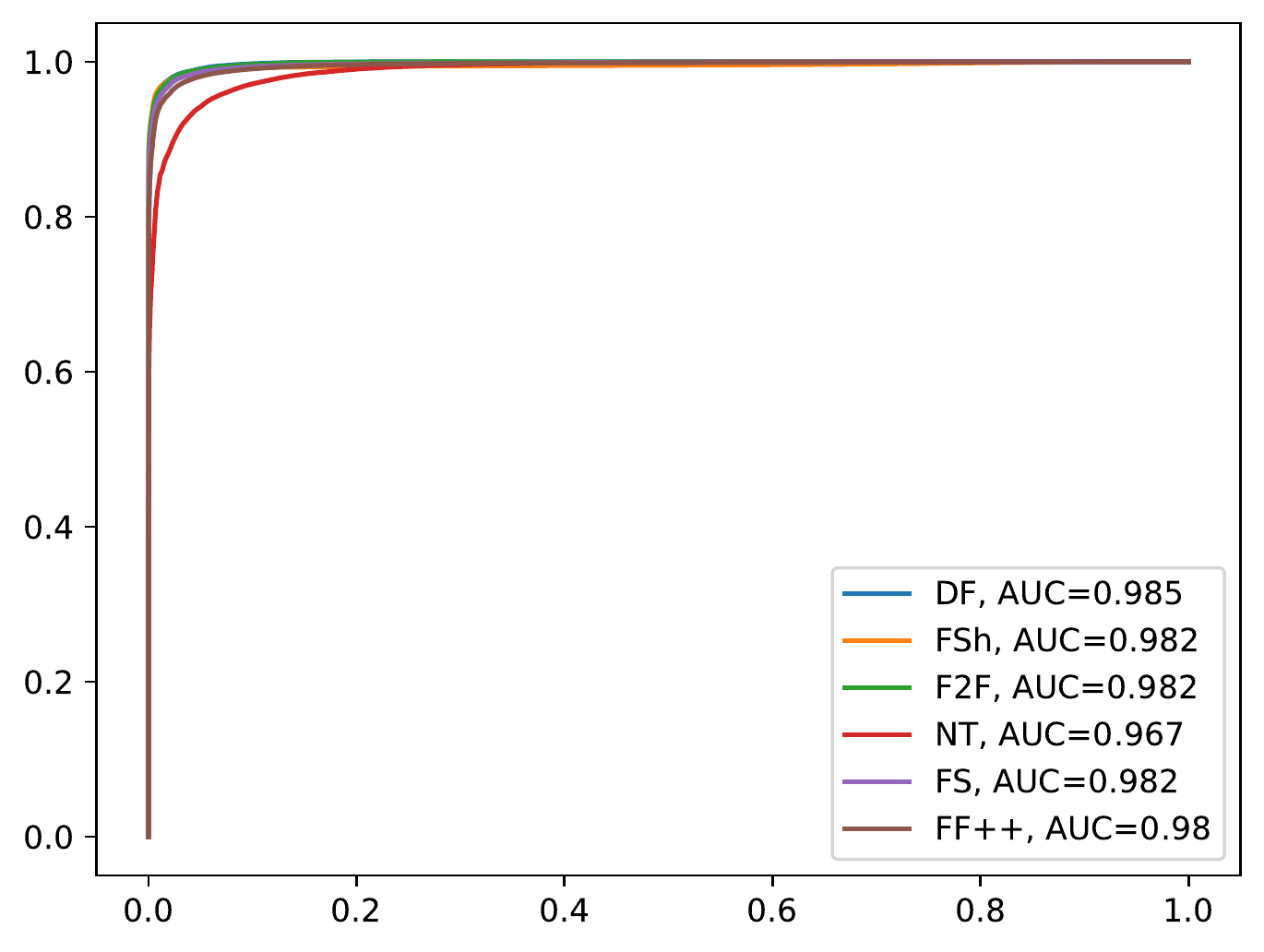}
    \caption{ROC curve for video level results. Model was trained on FaceForensics++ (FF++) and tested on the five sub-datasets within FF++ and all of FF++. This is an illustration of a part of Table 3 in the main paper.}
\end{figure}

\textbf{IoU for random guessing algorithm}

Let the ground truth map be $GT_{map}$ and predicted segmentation map be $P_{map}$. Both will be 1-D vectors of equal length with a predicted Boolean class ($R$ or $F$) for each frame in the video.
\begin{equation}
  GT_{map} = \{ R R R R R R F F F R R ... \}
\end{equation}
\begin{equation}
  P_{map} = \{ R R R R R R F F F R R ... \}
\end{equation}
\begin{equation}
    IoU = \frac{Intersection}{Union}
    = \frac{|GT_{map} \cap P_{map}|}{| GT_{map} \cup P_{map} |}
\end{equation}

Observation: ${|GT_{map} \cap P_{map}|}$ is the count of correctly predicted frames, and $| GT_{map} \cup P_{map} |$ is the count of correctly predicted frames and wrongly predicted frames $\times 2$. 

IoU falls in the range $[0, 1]$; where the greater the value, the better the predicted segment map. 
Although the theoretical lower bound of IoU is zero, in practice it is useful to understand how a random guessing algorithm will be scored. 
Let $f$ be the ratio of Real frames in the $GT_{map}$ and $p$ be the probability at which the randomly predicted frame in $P_{map}$ is classified as Real. The graph below shows the possible ${|GT_{map} \cap P_{map}|}$ values (call it $S$).

\tikzstyle{level 1}=[level distance=20mm, sibling distance=20mm]
\tikzstyle{level 2}=[level distance=30mm, sibling distance=10mm]
\tikzstyle{level 3}=[level distance=20mm]

\begin{tikzpicture}[grow=right,->]
\begin{scope}[yshift=-6cm]
  \node {$.$}
    child {node {$Actual Real$} 
      child {node {$Pred Real$}
        child[-] {node{$1$}}  
      edge from parent node [below] {$1-p$}
      }
      child {node {$Pred Fake$}
        child[-] {node{$0$}}  
      edge from parent node [below] {$p$}
      }
    edge from parent node [below] {$1-f$}
    }
    child {node {$Actual Fake$}
      child {node {$Pred Real$}
        child[-] {node{$0$}}
      edge from parent node [below] {$1-p$}
      }
      child {node {$Pred Fake$}
        child[-] {node{$1$}}  
      edge from parent node [below] {$p$}
      }
    edge from parent node [below] {$f$}
    };
\end{scope}
\end{tikzpicture}

For a single frame, the expected value of $S$ is,
\begin{equation}
\begin{aligned}
    E(S) {} &= f.p.1 + f(1-p).0 + (1-f).p.0 + (1-f).(1-p).1 \\
            &= 1 + 2.f.p - f - p = \alpha
\end{aligned}
\end{equation}

For $T$ total frames $E(S)=T\alpha$. Using our observation above $| GT_{map} \cup P_{map} | = 2(T-T\alpha)$. Therefore IoU can be calculated as,

\begin{equation}
\begin{aligned}
    E(S) {} &= \frac{T\alpha}{T(2-\alpha)} \\
            &= \frac{1 + 2.f.p - f - p}{1 - 2.f.p + f + p}
\end{aligned}
\end{equation}

For a random guessing algorithm with probability $p=0.5$ for each class in a binary classification problem we have $IoU = 1/3$. This will be the random guessing baseline for IoU in our context. from equation (5).

\textbf{Smoothing Algorithm}

\begin{figure}[!ht]
    \centering
    \includegraphics[width=0.49\textwidth]{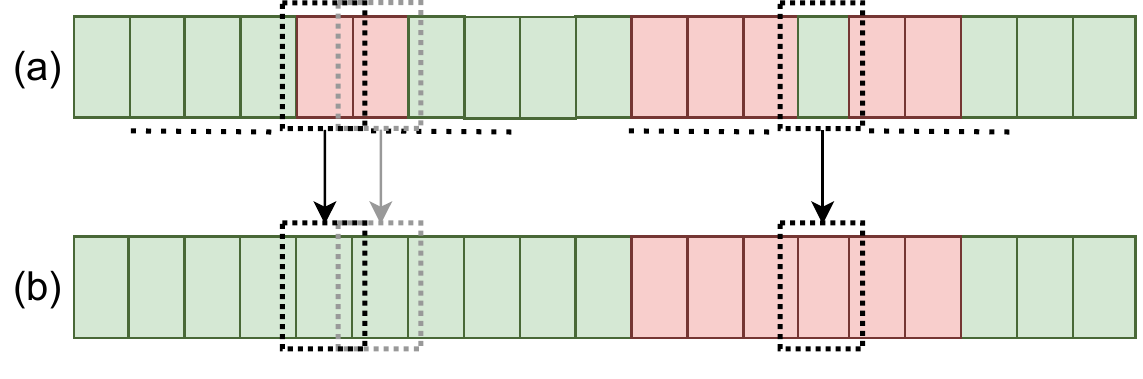}
    \caption{This figure depicts the visualization of our proposed approach for smoothing out noisy predictions. The first image (a) illustrates the raw frame-level predictions for a video, while the second image (b) shows the output after applying Algorithm \ref{algo:smoothing}. Green frames indicating \textcolor{darkgreen}{real} and red indicating \textcolor{red}{real} prediction. Frames can get their prediction changed based on the majority vote from past (left) and future (right) predictions, indicated by the dotted lines.}
  \label{fig:smoothing}
\end{figure}

The predictions of the ViT for the videos are frame-level and therefore there are often some noisy predictions. These noisy predictions can be corrected (Figure \ref{fig:smoothing}) with a simple smoothing technique. We have used Algorithm \ref{algo:smoothing} to smooth out noisy frame level prediction. In this algorithm a minimum fake-segment duration (in number of frames) is set. For each frame-prediction, majority voting is taken from predictions of past frames (on the left) and from future frames (on the right), and this helps determining the final label of that frame. Smoothing noisy predictions aids in better performance as can be seen in Table 6 in the main paper.

\begin{algorithm}
\caption{Smoothing noisy predictions.}\label{algo:smoothing}
    \begin{algorithmic}
    \Require $\rho$, the list of predictions per frame
    \Require $k \geq 0$, the offset
    \For{$i \leftarrow 0 \dots len(\rho)$}
    \State $\rho_{left} \gets$ sub-list of size $k$ on left of $\rho[i]$
    \State $\rho_{right} \gets$ sub-list of size $k$ on right of $\rho[i]$
    \State $M_{left} \gets$ majority-vote$(\rho_{left})$
    \State $M_{right} \gets$ majority-vote$(\rho_{right})$
    \If{$\rho_{left}$ is empty $~ and ~ \rho[i] \neq M_{right}$}
        \State $\rho[i] \gets M_{right}$
    \ElsIf{$\rho_{right}$ is empty $~ and ~ \rho[i] \neq M_{left}$}
        \State $\rho[i] \gets M_{left}$
    \ElsIf{$M_{left} = M_{right} ~ and ~ \rho[i] \neq M_{left}$}
        \State $\rho[i] \gets M_{left}$
    \EndIf
    \EndFor
    \State \Return $\rho$
    \end{algorithmic}
\end{algorithm}

\begin{figure}[!h]
    \centering
         \includegraphics[width=0.49\textwidth]{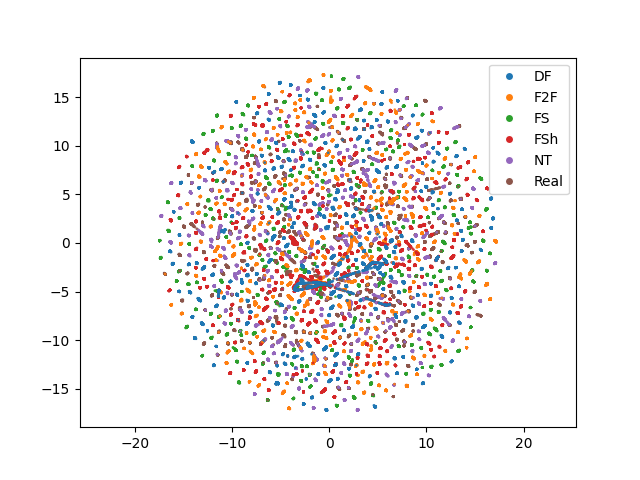}
    \caption{Uniform Manifold Approximation \& Projection (UMAP) on the spatial embeddings (from ViT) on the sub-datasets in FF++.}
    \label{fig:enter-label}
\end{figure}


\begin{figure}[!h]
     \centering
     \begin{subfigure}[b]{0.49\textwidth}
         \centering
         \includegraphics[width=\textwidth]{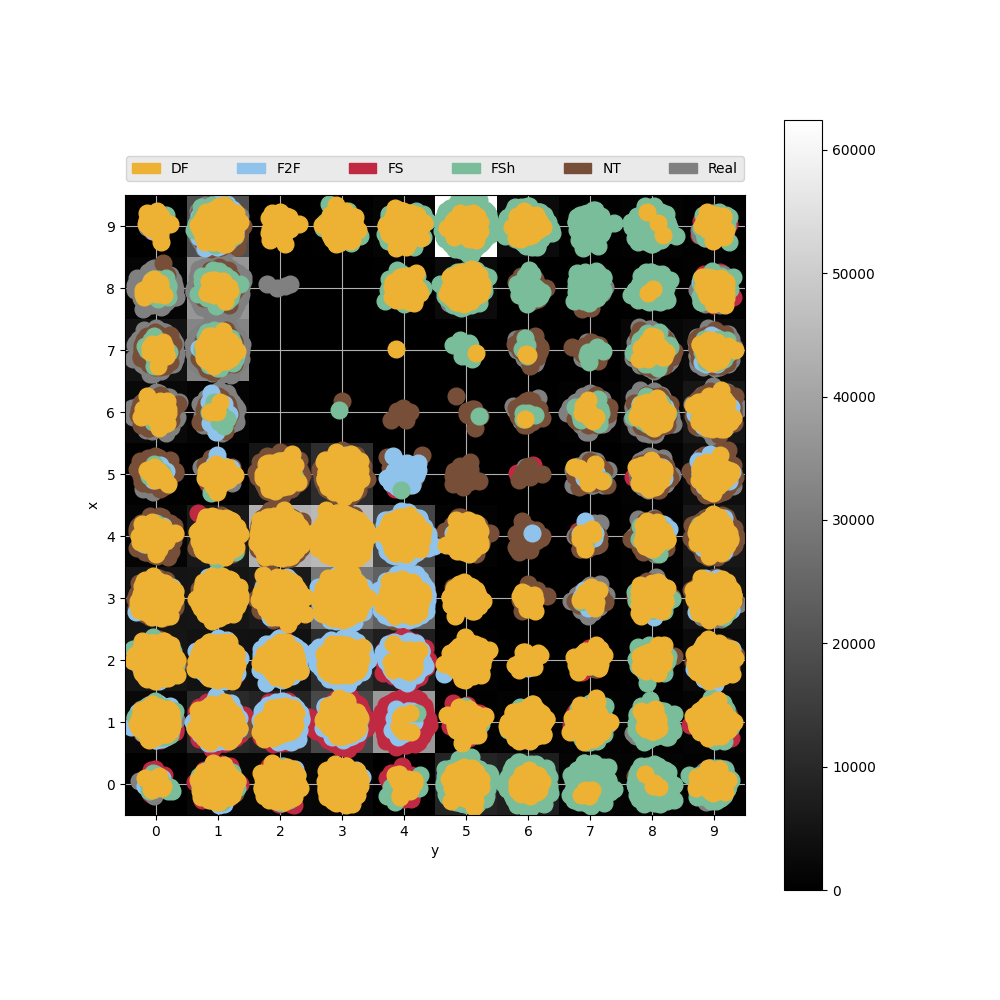}
         \caption{Self-organizing map (SOM)}
         \label{fig:som}
     \end{subfigure}
     \begin{subfigure}[b]{0.49\textwidth}
         \centering
         \includegraphics[width=\textwidth]{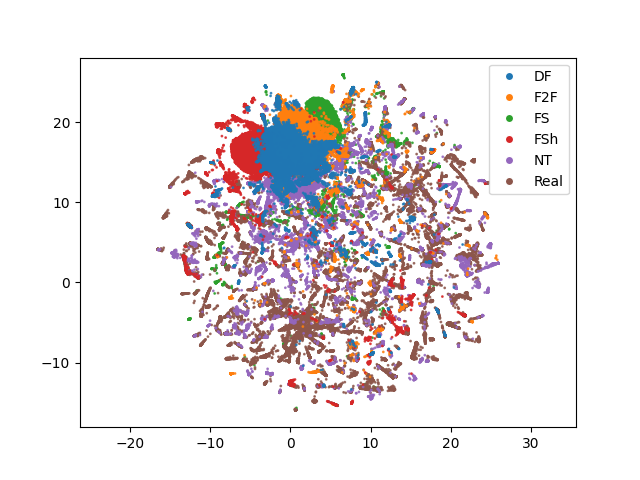}
         \caption{Self-Organizing Nebulous Growths (SONG)}
         \label{fig:song}
     \end{subfigure}
    \begin{subfigure}[b]{0.49\textwidth}
         \centering
         \includegraphics[width=\textwidth]{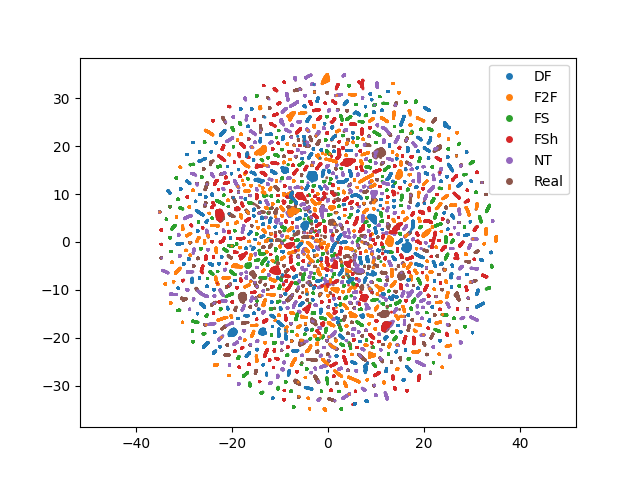}
         \caption{t-distributed Stochastic Neighbor Embedding (t-SNE)}
         \label{fig:tsne}
     \end{subfigure}
        \caption{Visualizations on the spatial embeddings (from ViT) on the sub-datasets in FF++.}
        \label{fig:visualizations_1}
\end{figure}

\end{document}